\documentclass[10pt,twocolumn,letterpaper]{article}

\usepackage[pagenumbers]{wacv} 

\usepackage{url}
\usepackage{array}
\usepackage{bm}
\usepackage{xfrac}
\usepackage{mathrsfs}
\usepackage{latexsym}
\usepackage{xcolor}
\usepackage{ulem}
\usepackage{stfloats}
\usepackage{float}
\usepackage{pdflscape}
\usepackage{colortbl}

\usepackage{enumitem}
\usepackage{graphicx}
\usepackage{amsmath}
\usepackage{amssymb}
\usepackage{multirow}
\usepackage{booktabs}
\usepackage{ctable}
\usepackage{xtab}
\usepackage{rotating}
\usepackage{threeparttable}

\usepackage[lined,ruled]{algorithm2e}

\usepackage[pagebackref,breaklinks,colorlinks]{hyperref}

\def\mask{\mathbf{m}}

\let\emph\textit

\newcolumntype{C}[1]{>{\centering\arraybackslash}p{#1}}
\newcolumntype{L}[1]{>{\raggedright\arraybackslash}p{#1}}
\newcolumntype{R}[1]{>{\raggedleft\arraybackslash}p{#1}}

\usepackage[capitalize]{cleveref}
\crefname{section}{Sec.}{Secs.}
\Crefname{section}{Section}{Sections}
\Crefname{table}{Table}{Tables}
\crefname{table}{Tab.}{Tabs.}

\def\wacvPaperID{651} 
\def\confName{WACV}
\def\confYear{2024}

\begin{document}

\title{Assist Is Just as Important as the Goal: Image Resurfacing to Aid Model's Robust Prediction}

\author{Abhijith Sharma\\
University of British Columbia\\
BC, Canada\\
{\tt\small sharma86@student.ubc.ca}
\and
Phil Munz\\
TrojAI Inc.\\
NB, Canada\\
{\tt\small phil.munz@troj.ai}
\and
Apurva Narayan\\
Western University\\
ON, Canada\\
{\tt\small apurva.narayan@uwo.ca}
}
\maketitle

\begin{abstract}
Adversarial patches threaten visual AI models in the real world. The number of patches in a patch attack is variable and determines the attack's potency in a specific environment. Most existing defenses assume a single patch in the scene, and the multiple patch scenarios are shown to overcome them. This paper presents a model-agnostic defense against patch attacks based on total variation for image resurfacing (TVR). The TVR is an image-cleansing method that processes images to remove probable adversarial regions. TVR can be utilized solely or augmented with a defended model, providing multi-level security for robust prediction. TVR nullifies the influence of patches in a single image scan with no prior assumption on the number of patches in the scene. We validate TVR on the ImageNet-Patch benchmark dataset and with real-world physical objects, demonstrating its ability to mitigate patch attack.   
\end{abstract}

\section{Introduction}
\label{sec:intro}

The vulnerabilities of CNNs against adversarial corruption are widely known \cite{akhtar2018threat}. Adversaries in the physical world can have a devastating impact, endangering human lives \cite{athalye2018synthesizing}, \cite{eykholt2018robust}, \cite{ren2021adversarial}. Physical corruption can be natural (like snow, dust, or blur) or artificially designed by malicious attackers (adversarial attacks). Natural noises are less threatening, and robustness against them is extensively explored \cite{rebuffi2021data}, \cite{shorten2019survey}. Brown et al.'s proposal of the adversarial patch \cite{brown2017adversarial} fueled the research on physical-world attacks. An adversarial patch is an overt, optimally formulated, and localized perturbation, printed as a poster in the scene \cite{sharma2022adversarial}, \cite{wei2022physical}. Although initial works only considered a single patch \cite{karmon2018lavan}, \cite{liu2018dpatch}, \cite{liu2019perceptual}, multiple patches provide more freedom to the attacker to design a stronger adversary \cite{sharma2023vulnerability}. 

CNNs have made remarkable progress in making human-like predictions \cite{mnih2015human}, \cite{taigman5closing}, yet, their failure against adversarial patches concerns their real-world deployment. Hence, the focus has been on defending CNN models in recent years. A CNN model that is either trained to be robust \cite{rao2020adversarial}, \cite{metzen2021meta} or augmented with a defensive technique \cite{mccoyd2020minority}, \cite{xiang2022patchcleanser}, will be able to protect itself from adversaries. Most existing defenses require access to the CNN's parameters directly or indirectly \cite{wei2022physical}. However, accessing model weights is sometimes infeasible due to privacy concerns and expensive model training limits the defense's applicability.

 A typical patch attack involves the manipulation of the scene rather than the model itself. Hence, is there a way to eradicate corruption at the root level by cleansing unwanted and suspicious patterns from the image? As the title suggests:``assist is just as important as the goal" we propose image resurfacing to assist the goal of the model's robust prediction. This work acts as a first line of defense against adversarially patched images. It can be augmented with any CNN model effortlessly. The motivation is not to substitute the robust models but to implement multi-level security. Even in cases where designing robust models are complicated, this technique can even independently mitigate the influence of adversarial patches from the scene.
\section{Related Work}
Designing a defense for patch attacks is a location identification problem. The detected patch's location can be masked or inpainted to mitigate the adversarial influence. Initial defenses primarily utilize a saliency map to locate adversarial patches and mask them \cite{hayes2018visible}, \cite{chou2020sentinet}, \cite{naseer2019local}. The method is simple, but the performance depends on the quality of the saliency map. Generating a salience map also requires access to the model's weights, which might be restricted sometimes. Moreover, detecting multi-patch attacks is a challenge due to multiple salient regions. The next set of defenses proposes a particular CNN architecture based on the small receptive field \cite{xiang2020patchguard}, \cite{xiang2021patchguard++}. A smaller field leads to the reduced global influence of a localized patch. The major drawback is the dependence on special robust architectures with relatively low clean accuracy. 

There have been proposals of defenses based on adversarial training, but they are unscalable due to the nuance of the patch attack \cite{rao2020adversarial}, \cite{gittings2020vax}. The methodology of adversarial training against digital attacks like PGD and FGSM cannot directly translate to patch attacks. On the other hand, certified defense is ideal as it provides a confidence guarantee along with their defending properties \cite{chiang2020certified}, \cite{levine2020randomized}, \cite{xiang2022patchcleanser}. However, such defenses come with an additional cost of computational complexity and are slower than their counterparts. Also, no existing defense could formulate guarantees when multiple patches are in the scene. Certain architectures show inherent robustness against adversarial patches \cite{chen2022towards}, \cite{salman2022certified}, \cite{li2022vip}, yet their performance evaluation against multi-patch attacks is still in progress. Overall, there has been considerable work on single patch-based defenses, and defending against multi-patch attacks has been challenging.

\begin{figure*}
     \centering 
     \includegraphics[scale=0.12]{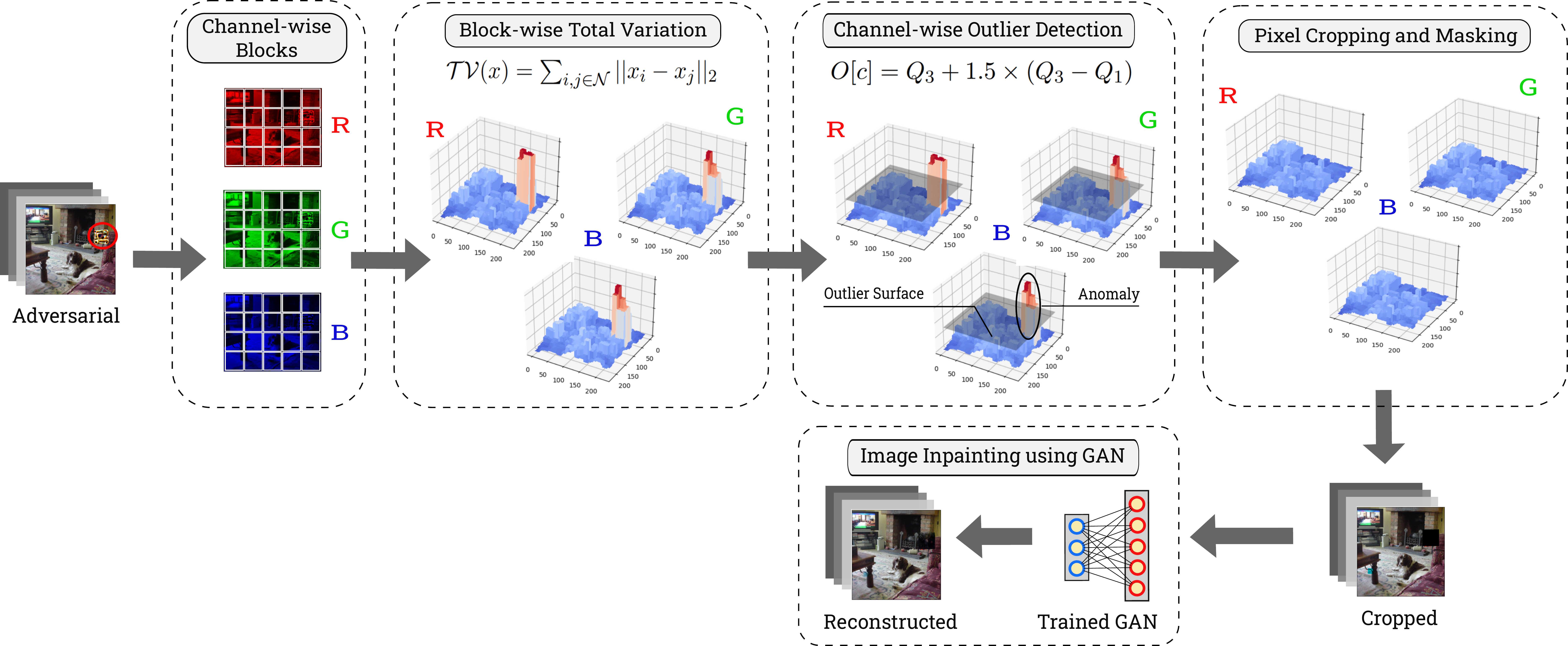}
     \caption{Framework of the total-variation based outlier detection followed by GAN inpainting to carry out image resurfacing.}
     \label{fig:tvr_frame}
\end{figure*}

\section{Our Contribution}

\begin{figure}[t]
     \centering
     \includegraphics[scale = 0.11]{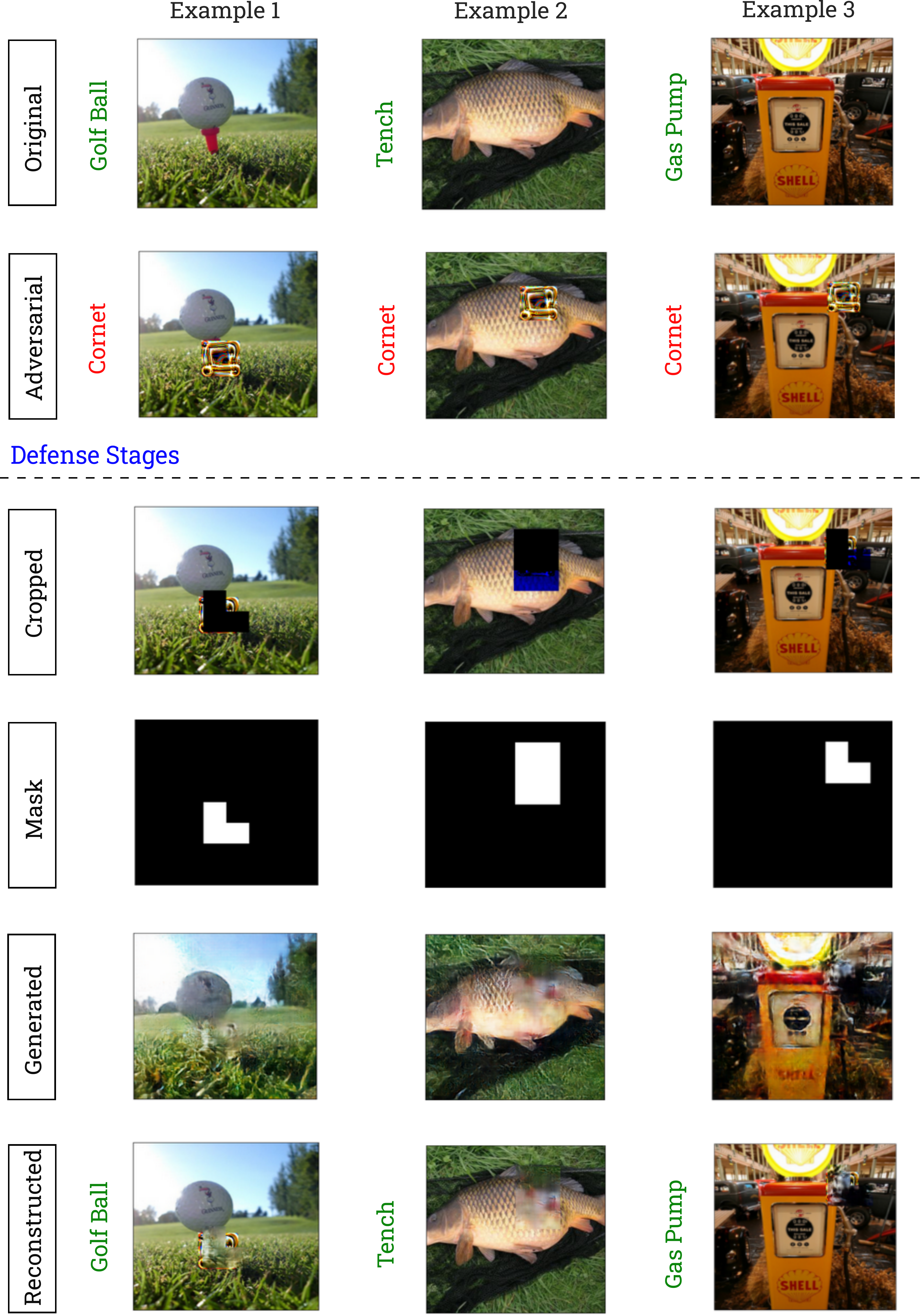}
     \caption{Illustration of TVR Demonstration with three examples.}
     \label{fig:def_demo}
\end{figure}

In this work, we utilize a popular image processing technique called total variation (TV) \cite{chambolle2010introduction}, \cite{sharif2016accessorize} to perform image resurfacing, and call it total-variation based image resurfacing (TVR). The steps involved in the TVR is shown in Figure \ref{fig:tvr_frame}. The resurfaced image is close to its original form such that even a normally trained model can perform robust prediction to some extent. The TV score has been previously utilized in adversarial defenses to understand pixel's spatial complexity in an image \cite{siddiqui2021novel}, \cite{guo2017countering}. Figure \ref{fig:def_demo} demonstrates how an adversarial image is processed and reconstructed to mitigate the adversarial patch under three instances.

The TVR is specifically designed to work against adversarial patches as it exploits the properties of patch attacks like localised and contiguous perturbation of pixels. But more importantly, the perturbation level of patch attacks are higher than that of digital adversarial attacks, which primarily inspires the development of TVR. The TVR only works on the image with no need of model or task information, which extends it's applicability to numerous applications. The characteristics of TVR are summarised below:
\begin{itemize}
\setlength\itemsep{-0.2em}
    \item \textbf{Model Agnostic: }The performance of the TVR is independent of the CNN model architecture.  
    \item \textbf{Patch Number Agnostic: } Can overcome multi-patch attack with no assumption on the number of patches. 
    \item \textbf{Patch Location Agnostic: } The patches will be detected irrespective of their location in the scene.
    \item \textbf{Location detection: } TVR is capable of determining approximate location of patches in the scene. 
\end{itemize}

\section{TVR Formulation}
\label{sec: tvr_form}

This section discusses the formulation of total-variation based image resurfacing (TVR) method. The framework in Figure \ref{fig:tvr_frame} illustrates image cleansing using TVR involving five broad steps as explained below:
\vspace{0.1cm}

\noindent \textbf{Channel-wise Blocks: } TVR calculates a channel-wise variation in pixel values, as an attack might not equally affect each channel. Averaging TV scores for three channels might lead to a loss of information. Hence, we perform a channel-wise inspection and later make image-level conclusions. The first step in TVR involves dividing the image into multiple equal-sized blocks as shown in Figure \ref{fig:tvr_frame}.

Converting an image into a block set is explained in Algorithm \ref{alg:img2blk}. The block set is a collection of equally sized blocks formed by dividing the image, as shown in Figure \ref{fig:tvr_frame}. We denote the block set as $\mathcal{B} \in \mathbb{R}^{3 \times \texttt{nk} \times k \times k}$, where $nk$ is the total number of blocks in the block-set, with each block being $k \times k$. The $\mathcal{B}$ stores channel-wise information. The input to the Algorithm is the image $\bm{x}$ and block-size $k$, and the output is the block-set $\mathcal{B}$.
\begin{algorithm}[b]
\LinesNumbered
\caption{\textsc{IMAGE\_TO\_BLOCK}}
\label{alg:img2blk}
\KwIn{\small Image $\bm{x} \in {\mathbb{R}}^{3 \times n \times n}$,  Block-size $k \times k$.}
\KwOut{\small  Block-set $\mathcal{B} \in \mathbb{R}^{3 \times \texttt{nk} \times k \times k}$}

$\texttt{nk} = (n\times n)/(k\times k) $ \hspace{0.05cm} // \texttt{\scriptsize number of blocks in image}

$\mathcal{B} \in \mathbb{R}^{3 \times \texttt{nk} \times k \times k}$ \hspace{0.3cm} // \texttt{\scriptsize block-set dimensions}

$\texttt{nrow} = (n / k) $ \hspace{0.3cm} // \texttt{\scriptsize number of blocks in each row}

\For{$ c \leftarrow 0,1,2$}{ 
\For{$ b \leftarrow 0$ to $\texttt{nk}$}{
\For{$ i \leftarrow 0$ to $k$}{
\For{$ j \leftarrow 0$ to $k$}{
$\mathcal{B}[c][b][i][j] = \bm{x}[c][\texttt{int}(b / \texttt{nrow})\times k+i][(b \% \texttt{nrow}) \times k+j]$
}}}}

\end{algorithm}

\noindent \textbf{Block-wise Total variation: }This step calculates the block-wise total variation (TV) score. The TV is a metric to determine the complexity of an image to its spatial variation in pixel values. TV score has been extensively used in various applications in image processing literature. The TV of an RGB image $\bm{x} \in {\mathbb{R}}^{3 \times n \times n}  $ is defined as follows:

\begin{equation}
    \mathcal{TV}(\bm{x}) = \sum_{i,j \in \mathcal{N}} ||x_i - x_j||_p^q
\end{equation}

where $\mathcal{N}$ is the pixel neighborhood. Typically, $\mathcal{N}$ consists of the horizontally and vertically adjacent pixels, and $||.||_p^q$ is the $l_p$ norm to the power of $q$. This work utilizes $\mathcal{TV}$ with $l_2$ norm and $q=1$ (refer Line 3, Algorithm \ref{alg:tvd}). 

Figure \ref{fig:tvr_frame} shows 3-D surface plots of the TV score over the image landscape (block set) for each channel. The neighborhood $\mathcal{N}$ corresponds to each block $b$ in the block set $\mathcal{B}$. Hence, the number of data points forming the plot's surface is the total number of blocks $nk$ that the image is divided into. The red mountain peaks translate to the highly probable region of adversarial perturbations from high TV score. Hence, the TV score based surface plots provides the approximate location of malicious patches in the scene.

\begin{algorithm}[t]
\LinesNumbered
\caption{TV-based Image Resurfacing (TVR)}
\label{alg:tvd}
\KwIn{\small Adversarial image $\bm{x'}$,  Cropped image $\bm{x_c}$, Generated image $\bm{x_g}$ ($\bm{x'}, \bm{x_c},\bm{x_g} \in {\mathbb{R}}^{3 \times n \times n}$), Block-size $k \times k$ , Mask set $\mathcal{M} \in {\mathbb{R}}^{3 \times \texttt{nk} \times k \times k}$, , Generator $\mathcal{G}$(.), Percentile function $\mathcal{P}^{th}$(.)}

\KwOut{\small Reconstructed image $\bm{x}_{r} \in {\mathbb{R}}^{3 \times n \times n} $}

$\texttt{nk} = (n \times n)/(k \times k) $ \hspace{0.3cm} // \texttt{\scriptsize total number of blocks} 

$\mathcal{B} = \texttt{IMAGE\_TO\_BLOCK}(\bm{x}', k)$ \hspace{0.1cm} // \texttt{\scriptsize create block-set}

$\mathcal{TV}(x) = \sum_{i,j \in \mathcal{N}} ||x_i - x_j||_2$ \hspace{0.1cm} // \texttt{\scriptsize TV loss function} 
\For{$ c \leftarrow 0,1,2$}{
\For{$ b \leftarrow 0$ to $\texttt{nk}$}{

$tv[c][b] = \mathcal{TV}(\mathcal{B}[c][b])$ \hspace{0.2cm} // \texttt{\scriptsize TV loss}
}
$Q_1 = \mathcal{P}^{th}(tv[c], 0.25)$ \hspace{0.3cm} // \texttt{\scriptsize first quartile}

$Q_3 = \mathcal{P}^{th}(tv[c], 0.75)$ \hspace{0.3cm} // \texttt{\scriptsize third quartile}

$O[c] = Q_3 + 1.5 \times (Q_3- Q_1)$ \hspace{0.2cm} // \texttt{\scriptsize outliers}
}

\For{$ c \leftarrow 0,1,2$}{
\For{$ b \leftarrow 0$ to $nk$}{
\eIf{$O[c] > \mathcal{TV}(\mathcal{B}[c][b])$}{ 
$\mathcal{B}[c][b] \leftarrow 0$ 
      
$\mathcal{M}[c][b] \leftarrow 1$ 
}{
$\mathcal{M}[c][b] \leftarrow 0$ 
}
}}

$\bm{x_c} = \texttt{IMAGE\_TO\_BLOCK}^{-1}(\mathcal{B})$ \hspace{0.1cm} // \texttt{\scriptsize create image}

$\mask = \texttt{IMAGE\_TO\_BLOCK}^{-1}(\mathcal{M})$  \hspace{0.1cm} // \texttt{\scriptsize create mask}

$\bm{x}_g$ = $\mathcal{G}$( $\bm{x}_c$ ) \hspace{0.1cm} // \texttt{\scriptsize generated image using GAN}

$\bm{x}_{r} = (1-\mask) \odot \bm{x}_c + \mask \odot \bm{x}_g$ \hspace{0.1cm} // \texttt{\scriptsize final image} 
\end{algorithm}

\vspace{0.1cm}
\noindent \textbf{Channel-wise Outlier Detection: }This step calculates the outlier $O$ using the simple formula $Q_3 + 1.5 \times (Q_3- Q_1)$. The $Q1$ and $Q3$ is the first ($25^{th}$ percentile) and third quartile ($75^{th}$ percentile) of the TV distribution, respectively. 
Although outliers can be present on both ends of the spectrum, but the outliers on the higher end corresponds to the high total variation score which could have caused by patch attacks. Since the TV score will vary for each channel, so will the outliers. Hence, we calculate channel-wise outliers in this block (refer Lines (4-11), Algorithm \ref{alg:tvd}).

\vspace{0.1cm}
\noindent \textbf{Pixel Masking: } The outliers from the previous step are used to mask the pixels of the blocks $b$ where the values are larger than the outlier, indicating the probability of patch presence. For pixels whose values are larger than the outlier, then the entire block's pixel values are assigned zero to form a cropped image $x_c$. For the final reconstructed image, a mask $m$, which is created out of mask set $\mathcal{M} \in {\mathbb{R}}^{3 \times \texttt{nk} \times k \times k}$ similar to that of image's block set. However, unlike the image, the mask is a binary tensor, where $m \in \{0,1\}^{3 \times n \times n}$. 

The pixel values of all blocks in the mask set are zero by default. The values are set to one for only those blocks in the mask set which correspond to the suspicious blocks (tagged by outlier detection) in the image's block set (refer Lines (13-24), Algorithm \ref{alg:tvd}). The algorithm of converting block set into an image and mask set into a mask is inverse of Algorithm \ref{alg:img2blk} ($\texttt{IMAGE\_TO\_BLOCK}$), which we call $\texttt{BLOCK\_TO\_IMAGE} = \texttt{IMAGE\_TO\_BLOCK}^{-1}$. Specifically, the operands on Line 8, Algorithm \ref{alg:img2blk} would interchange. The input to $\texttt{IMAGE\_TO\_BLOCK}^{-1}$ is the block set $\mathcal{B}$ and output is the RGB image $\bm{x}$. 







\vspace{0.1cm}
\noindent \textbf{Image Reconstruction: }This step concerns reconstructing or inpainting the cropped image $x_c$ using a generative adversarial network (GAN). From the previous step, we obtain a mask $m$ and a cropped image $x_c$. The cropped image has masked regions as pixel values are assigned zero for the suspicious blocks. Although the previous step is sufficient to mitigate the influence of the patch, we lose some information due to masking, which degrades the prediction accuracy. The image inpainting helps to reduce the gap between natural and adversarial accuracy. 

A crucial component of this step is the quality of images produced by generator. The generator $\mathcal{G}$ model is an autoencoder which tries to produce fake images to fool the discriminator $\mathcal{D}$. The discriminator is basically a CNN model followed by a Sigmoid function that finally gives a single scalar as output, whether the input image is real or fake. The loss function of the generator $\mathcal{G}$ consists of two parts:

\vspace{0.1cm}
\emph{Reconstruction Loss: } The loss of recreating an image using generator $\mathcal{G}$. It assists in capturing the structure and coherence of the missing region based on its context. The reconstruction loss is a $l_2$ loss, which reduces the mean pixel wise error leading to a blurry image. Hence, it is not sufficient to use reconstruction loss. The loss is defined as:

\begin{equation}
    \mathcal{L}_{reconstruct} = \mathbb{E}_{\hat{x} \sim P_c} \left[ \frac{1}{N} \sum_{i=1}^{N} {||\mathcal{G}(\hat{x}^{(i)}) - \hat{x}^{(i)}||}^2 \right]
\end{equation}

where $\hat{x}$ is the image to be reconstructed ($x_c$ in the Algorithm \ref{alg:tvd}) and $\mathbb{P}_c$ is the distribution of all $x_c$.

\vspace{0.1cm}
\emph{Adversarial Loss: } With the reconstruction loss we have blurry image and adversarial loss adds some realism to offset the blur. The goal of a generator in a GAN is to produce realistic images to fool the discriminator. The adversarial loss measures the extent to which generator is able to mislead the discriminator. This loss is defined as: 

\begin{equation}
\begin{aligned}
    \mathcal{L}_{adver} = \max_{\mathcal{D}} \hspace{0.2cm}  \mathbb{E}_{x \sim P_r} \left[ \log (\mathcal{D}(x)) \right] + \\
    \mathbb{E}_{\hat{x} \sim P_c} \left[ \log (1-\mathcal{D}(\mathcal{G}(\hat{x})))\right]
\end{aligned}
\end{equation}

where $x$ is the real/original image and $\mathbb{P}_c$ is the distribution of real images. The adversarial loss is augmented with reconstruction loss for image inpainting. The augmentation is done using a hyper-parameter $\lambda$. Typically we use $\lambda_{adver} + \lambda_{reconstruct} = 1$. The adversarial network produces a realistic image. The total loss function is given:
\begin{equation}
    \mathcal{L}_{total} = \lambda_{adver} \mathcal{L}_{adver} + \lambda_{reconstruct} \mathcal{L}_{reconstruct}
\end{equation}
To summarise, the aim is to minimize the $\mathcal{L}_{total}$. Hence, the min-max optimization as a apart of $\mathcal{L}_{adver}$ ensures that the generated images look similar to real ones, and minimization of $\mathcal{L}_{reconstruct}$ assists generator in this process. The new image $\bm{x}_{new}$ is reconstructed with the $\bm{x_c}$ and $\bm{x_g}$ using the formula $(1-\mask) \odot \bm{x}_c + \mask \odot \bm{x}_g$ as stated in Line 26, Algorithm \ref{alg:tvd}. The $\odot$ is a Hadamard operator representing element-wise multiplication between the tensor operands.


\section{Experiments and Results}
\label{sec: tvr_exp}

This section examines TVR's defending ability over a set of experiments. We use two datasets, one is ImageNet-Patch benchmark dataset \cite{pintor2023imagenet}, which is a dataset made with 10 adversarially trained patches (Soap Dispenser (SD), Cornet (CN), Plate (P), Banana (B), Cup (C), Typewriter  (TK), Electric Guitar (EG), Hair Spray (HS), Socks (S), Cellphone (CT)) over the ImageNet \cite{krizhevsky2017imagenet} dataset. The Image-Net Patch dataset only consists of single patches, and to evaluate our method against the most potent design of multi-patch attack \cite{sharma2023vulnerability} we use Imagenette dataset. The Imagenette is a 10 class subset of ImageNet with 10,000 images shared almost equally between 10 classes. The adversarial patch used with this dataset is self trained for the target class of `Golf Ball'.

\subsection{Experimental Setup}
\label{subsec:expsetup}
The patches in the ImageNet-Patch dataset are trained with ensemble training on AlexNet \cite{krizhevsky2017imagenet}, ResNet18 \cite{he2016deep}, and SqueezeNet \cite{iandola2016squeezenet}. On the other hand, VGG16 \cite{simonyan2014very}, GoogleNet \cite{szegedy2015going} and Inception v3 \cite{szegedy2016rethinking} is used for black-box testing. All models are off-the-shelf from PyTorch's Torchvision library. For Imagenette, we used VGG16 and ResNet18 that are pre-trained from Torchvision and fine-tuned on Imagenette. The patches are trained in a white-box testing. All images are resized to 224 $\times$ 224 for consistency in the results. The patch size in the ImageNet-Patch dataset is approximately 5-6\%  of the image size. Imagenette's patch size varies between 4\%, 8\%, and 12\% of the image size.  The self-trained single and multiple patches over Imagenette involved patch training over 8,000 images from the train set. The patches trained over Imagenette are less robust than those in the ImageNet-Patch benchmark. However, they are sufficient to validate the multi-patch attack results. Training patches for the Imagenette dataset is easier, as it is possible to attack with less potent adversarial patches due to the simplicity of the dataset.

The inpainting in TVR is done using a pre-trained generator. We use a rudimentary custom defined GAN architecture that is easy to train and just enough to demonstrate our framework. However, using a better and deeper generator will improve the inpainting process. The input and output of the generator have exact dimensions ${3\times224\times224}$. With the mean-squared loss, we used PyTorch's \texttt{Adam} optimizer with a learning rate of 0.002 over 300 epochs for generator training over a subset of the ImageNet train-set. For the TVR, the block size is chosen as 28 $\times$ 28, which we found is the most appropriate size for this dataset, as discussed in Section \ref{sec: blk_acc}. 
Quantitative experiments were carried out on the single Nvidia RTX A6000, and the resources available with Google Colab were computationally sufficient for visualizations and plots.

\subsection{Analysis of TVR Against Adversarial Patches}
We primarily use three evaluation metric: clean accuracy, adversarial accuracy and targeted success rate. We report all metrics for top-k $\mathcal{T}_k$ scenario, which is determined by checking if the original class $y$ is within the k highest predicted class (based on CNN's probabilistic output). In our experiments, since ImageNet-Patch dataset has 1000 classes, it is reasonable to calculate Top-1 and Top-3. However, for Imagenette, we limit ourselves to only Top-1 as it is just a 10 class subset. 

\vspace{0.2cm}
\noindent \textbf{Performance on ImageNet-Patch Benchmark: } The performance of TVR is averaged over 10 adversarial patches and is shown in Table \ref{tab: tvr_in}. The `Naive' model refers to the one without any defensive properties. The `Target' is the targeted success rate of the attack. We also report the average for the ensemble models separately to understand how TVR performs in a white-box setting separately. Since the patch is trained on models involved in the white-box ensemble set, the accuracy will be lower as the patch will be more effective. We observe a 42\% increase in top-1, a 34\% increase in top-3, and a 25\% increase in top-10 overall adversarial accuracy with TVR. 

\begin{table}[t]
\centering
\caption{\small
Performance of the TVR on benchmark ImageNet-Patch Dataset. All values are reported in \%.
}\label{tab: tvr_in}
{
\scalebox{.68}{
\begin{tabular}{cc|ccc|ccc}
\toprule
\multicolumn{2}{c|}{\multirow{2}{*}{\bf Model}} & \multicolumn{3}{c|}{\textbf{Naive}} & \multicolumn{3}{c}{\textbf{TVR}}\\
 & & Clean & Adversarial & Target & Clean & Adversarial & Target \\
\midrule
\midrule
\parbox[t]{2mm}{\multirow{6}{*}{\rotatebox[origin=c]{90}{\textbf{TOP - 1}}}} 
& AlexNet & 64.2 & 24.9 & 6.9 & 56.3 & 53.6 & 0 \\
& ResNet18 & 78.1 & 47.8 & 13.1 & 76.7 & 68 & 0 \\
& SqueezeNet & 56.2 & 33.6 & 18.4 & 50.8 & 46.2 & 0 \\
& VGG16 & 74.7 & 53.8 & 7.6 & 74.2 & 71.3 & 0 \\
& GoogleNet & 74.3 & 54.4 & 2.6 & 72.2 & 64.4 & 0 \\
& Inception v3 & 72.1 & 40 & 6.2 & 72.1 & 56.9 & 0 \\
\rowcolor{lightgray!60} & \textbf{Ensemble} & \textbf{66.1} & \color{red}{\textbf{35.4}} & \textbf{12.8} & \textbf{60.7} & \color{blue}{\textbf{55.9}} & \textbf{0} \\
\rowcolor{lightgray!60} & \textbf{Overall} & \textbf{69.9} & \color{red}{\textbf{42.4}} & \textbf{9.0} & \textbf{66.7} & \color{blue}{\textbf{60.1}} & \textbf{0} \\
\midrule
\parbox[t]{2mm}{\multirow{6}{*}{\rotatebox[origin=c]{90}{\textbf{TOP - 3}}}}
& AlexNet & 86.5 & 39.8 & 14.4 & 86.0 & 75.6 & 0.2 \\
& ResNet18 & 96.5 & 69.1 & 25.3 & 96.3 & 91.1 & 0 \\
& SqueezeNet & 86.1 & 50.9 & 28.4 & 86.2 & 72 & 0.7 \\
& VGG16 & 96.6 & 71.3 & 12.9 & 94.5 & 89.8 & 0.2 \\
& GoogleNet & 94.9 & 74.4 & 6.9 & 94.8 & 84.9 & 0 \\
& Inception v3 & 92.0 & 62.7 & 13.6 & 88.1 & 81.8 & 0 \\
\rowcolor{lightgray!60} & \textbf{Ensemble} & \textbf{89.7} & \color{red}{\textbf{53.3}} & \textbf{22.7} & \textbf{89.3} & \color{blue}{\textbf{79.6}} & \textbf{0.3} \\
\rowcolor{lightgray!60} & \textbf{Overall} &  \textbf{92.1} & \color{red}{\textbf{61.4}} & \textbf{16.9} & \textbf{90.7} & \color{blue}{\textbf{82.5}} & \textbf{0.2}\\
\bottomrule
\end{tabular}
}}

\end{table}

The higher adversarial accuracy translates to superior robustness. With the TVR, we observe a marginal drop in clean accuracy for all the accuracy levels. It may be because of the false outlier predictions even in clean images due to the high total variation score in some regions in the scene. Moreover, the success rate of a targeted attack for the top-1 scenario with TVR is zero. Even for top-3, the success rate is extremely low compared to the Naive model. It demonstrates that TVR nullifies the influence of adversarial patches. Among the white-box models, the ResNet18 shows the highest level of robustness, with VGG16 being the most robust for the black-box setting.


\vspace{0.2cm}
\noindent \textbf{Performance against Multi-Patch attack on Imagenette: } For analyzing TVR on Imagenette, we use ResNet18 and VGG16, as shown in Table \ref{tab: tvr_it}. The patches for ResNet18 and VGG16 are trained individually using LaVAN \cite{karmon2018lavan} methodology and attacked in a white box setting. We use three scenarios with Imagenette: original image, image with single patch, and image with multi-patch. Moreover, we assume that a single patch of size $\delta$ is evaluated against the multi-patch attack (of $n$ patches) with the size of each patch being ($\delta / n$) for fair comparison. In the scope of this experiment, we use four patches for the multi-patch attack. 

Imagenette is a simple dataset, hence,  ResNet18 and VGG16 could achieve 99\% clean accuracy. For single patch attack, we see a drastic drop in adversarial accuracy of the Naive model from 4\% to 8\%, yet the TVR defended model is able to retain the accuracy. Even for a 12\% patch, the TVR can mitigate the patch's effect to a large extent. For the multi-patch, we notice that 4\% perturbation level is capable of reducing the accuracy to 7\%. For 8\% and 12\% patch size, the accuracy comes down to 0\%. Among ResNet18 and VGG16, we notice that VGG16 performs better with TVR, with accuracy above 95\% for all the perturbation levels and both attack types (except one - 12\% Multi-Patch attack).

\begin{table}[t]
\centering
\caption{\small
TVR's performance against Multi and Single patch attack on Imagenette for perturbations of 12\%, 8\% and 4\%, respectively.
}\label{tab: tvr_it}
{
\scalebox{0.69}{
\begin{tabular}{cc|c>{\columncolor[gray]{0.85}}c | c>{\columncolor[gray]{0.85}}c | c>{\columncolor[gray]{0.85}}c}
  \toprule
  \multicolumn{2}{c|}{\multirow{2}{*}{\bf Model}} & \multicolumn{2}{c|}{\textbf{Original}} & \multicolumn{2}{c|}{\textbf{Single Patch}} & \multicolumn{2}{c}{\textbf{Multi Patch}}\\
   & & Naive & TVR & Naive & TVR & Naive & TVR\\
  \midrule
  \midrule
  \parbox[t]{2mm}{\multirow{3}{*}{\rotatebox[origin=c]{90}{\textbf{4\%}}}} & ResNet18 & 99.7 & 98.5 & 88.7 & 95.8 & 1.1 & 96.3\\
  & VGG16 & 99.1 & 98.1 & 10.5 & 98.1 & 13.2 & 99.1\\
  & \textbf{Overall} &  \textbf{99.4} & \textbf{98.3} & \color{red}{\textbf{49.7}} & \color{blue}{\textbf{97.0}} & \color{red}{\textbf{7.1}} & \color{blue}{\textbf{97.1}}\\
  \midrule
  \parbox[t]{2mm}{\multirow{3}{*}{\rotatebox[origin=c]{90}{\textbf{8\%}}}}& ResNet18 & 99.7 & 96.2 & 20.3 & 94.1 & 0 & 73.5\\
  & VGG16 & 99.1 & 98.4 & 0 & 97.7 & 0 & 95.8\\
  & \textbf{Overall} &  \textbf{99.4} & \textbf{97.3} & \color{red}{\textbf{10.1}} & \color{blue}{\textbf{95.6}} & \color{red}{\textbf{0}} & \color{blue}{\textbf{84.6}}\\
  \midrule
  \parbox[t]{2mm}{\multirow{3}{*}{\rotatebox[origin=c]{90}{\textbf{12\%}}}}& ResNet18 & 99.7 & 97.3 & 9.6 & 87.4 & 0 & 69.3\\
  & VGG16 & 99.1 & 99.2 & 0 & 82.4 & 0 & 81.2\\
  & \textbf{Overall} &   \textbf{99.4} & \textbf{98.2} & \color{red}{\textbf{4.8}} & \color{blue}{\textbf{84.9}} & \color{red}{\textbf{0}} & \color{blue}{\textbf{75.2}}\\
  \bottomrule
\end{tabular}
}}
\end{table}

\vspace{0.2cm}
\noindent \textbf{Comparison with state-of-art defenses against Multi-Patch attack on ImageNet-Patch Benchmark: } In this study, we compare TVR against two model-agnostic defenses: Localized Gradient Smoothing (LGS) \cite{naseer2019local} and PatchCleanser (PC) \cite{xiang2022patchcleanser} along with a Naive (undefended) model for reference. The results are shown in Table \ref{tab: tvr_comp}. As the defenses are model-agnostic, we evaluate them for the six CNNs similar to Table \ref{tab: tvr_in}. We consider single and multiple patches (with one, two, three, and four patches) placed randomly over the scene. The size of each patch is around 5\% of the image area. The analysis is carried out over 100 randomly sampled images from the test set.

As evident from Table \ref{tab: tvr_comp}, even though PC is a certified defense, it does fail for multi-patch attacks. The PC is inherently designed for single patches and can be scaled to two patches by increasing the search complexity. However, TVR detects single and multiple patches in one image scan. The performance of LGS was similar to TVR for a single patch attack. However, TVR has a superior performance when it comes to multiple adversarial patches. 

For a two-patch attack (Multi-2), we observe that TVR outperforms LGS by 6\% and PC by 28\% on average across six models. Similarly, for a three-patch attack (Multi-3), the TVR outperforms LGS and PC by 12\% and 31\% , respectively. For four patches, we observe that all defended models have low accuracy because the scene occlusion is high and scene reconstruction is complex. But using advanced GAN for image inpainting can improve the performance of TVR substantially. However, TVR still outperforms LGS and PC in four patch attack scenarios, which proves TVR has the best patch mitigation technique overall.  

To summarize, Table \ref{tab: tvr_in}, \ref{tab: tvr_it} and \ref{tab: tvr_comp} showcases the ability of TVR to achieve robustness against single and multi-patch attacks on ImageNet dataset. We also did not notice a substantial difference between the white and black box settings. The primary reason is that the TVR works on the total-variation of pixel values, which has no connection with whether the pixels are trained in a white or black box fashion. Moreover, the TVR also shows superior performance against LGS and PC for multiple adversarial patches. 

\begin{table}[t]
\centering
\caption{\small
Comparison of existing defenses against TVR for single and multi-patch attack. The patch is of `banana' class. The values are the number of correct classification out of 100 test samples.
}\label{tab: tvr_comp}
\scalebox{.72}{
\begin{tabular}{cc|c|c|c|c|c}
\toprule
\multicolumn{2}{c|}{\textbf{Defense Method}} & \textbf{Clean} & \textbf{Single} & \textbf{Multi-2} & \textbf{Multi-3} & \textbf{Multi-4}\\
\midrule
\midrule
\parbox[t]{2mm}{\multirow{4}{*}{\rotatebox[origin=c]{90}{\textbf{AlexNet}}}} & Naive & \textbf{65} & 31 & 4 & 2 & 0 \\
& LGS \cite{naseer2019local} & 61 & \textbf{56} & 41 & 18 & 17\\
& PatchCleanser \cite{xiang2022patchcleanser} & 56 & 53 & 30 & 8 & 2 \\
& \textbf{TVR (ours)}& 56 & 56 & \textbf{40} & \textbf{27} & \textbf{20}\\
\midrule
\parbox[t]{2mm}{\multirow{4}{*}{\rotatebox[origin=c]{90}{\textbf{ResNet18}}}} & Naive & \textbf{78} & 57 & 5 & 0 & 0 \\
& LGS \cite{naseer2019local} & 68 & 65 & 56 & 22 & 4\\
& PatchCleanser \cite{xiang2022patchcleanser} & 74 & 46 & 16 & 4 & 1 \\
& \textbf{TVR (ours)} & 76 & \textbf{72} & \textbf{52} & \textbf{42} & \textbf{23}\\
\midrule
\parbox[t]{2mm}{\multirow{4}{*}{\rotatebox[origin=c]{90}{ 
\textbf{SqueezeNet}}}} & Naive 
& \textbf{57} & 18 & 0 & 0 & 0 \\
& LGS \cite{naseer2019local} &  52 & \textbf{48} & 30 & 9 & 4\\
& PatchCleanser \cite{xiang2022patchcleanser} & 54 & 20 & 4 & 3 & 0 \\
& \textbf{TVR (ours)}& 50 & \textbf{49} & \textbf{42} & \textbf{22} & \textbf{12}\\
\midrule
\parbox[t]{2mm}{\multirow{4}{*}{\rotatebox[origin=c]{90}{\textbf{VGG16}}}} & Naive & \textbf{75} & 39 & 4 & 4 & 0 \\
& LGS \cite{naseer2019local} & 73 & 64 & 42 & 24 & 11\\
& PatchCleanser \cite{xiang2022patchcleanser} & 75 & 44 & 17 & 6 & 2 \\
& \textbf{ TVR (ours)}& 74 & \textbf{66} & \textbf{46} & \textbf{43} & \textbf{24}\\
\midrule
\parbox[t]{2mm}{\multirow{4}{*}{\rotatebox[origin=c]{90}{\textbf{GoogleNet}}}} & Naive & \textbf{74} & 54 & 29 & 20 & 10 \\
& LGS \cite{naseer2019local} & 73 & 54 & 46 & 37 & 20\\
& PatchCleanser \cite{xiang2022patchcleanser} & 72 & 53 & 30 & 8 & 2 \\
& \textbf{ TVR (ours)}& 73 & \textbf{62} & \textbf{58} & \textbf{42} & \textbf{25}\\
\midrule
\parbox[t]{2mm}{\multirow{4}{*}{\rotatebox[origin=c]{90}{\textbf{Inception}}}} & Naive & \textbf{76} & 50 & 38 & 25 & 16 \\
& LGS \cite{naseer2019local} & 74 & 62 & 47 & 42 & 29\\
& PatchCleanser \cite{xiang2022patchcleanser} & 70 & 55 & 31 & 10 & 8 \\
& \textbf{TVR (ours)}& 72 & \textbf{65} & \textbf{60} & \textbf{48} & \textbf{33}\\
\bottomrule
\end{tabular}
}
\end{table}

\subsection{Visualization of TVR's Working}

\begin{figure}[t]
     \centering 
     \includegraphics[scale=0.115]{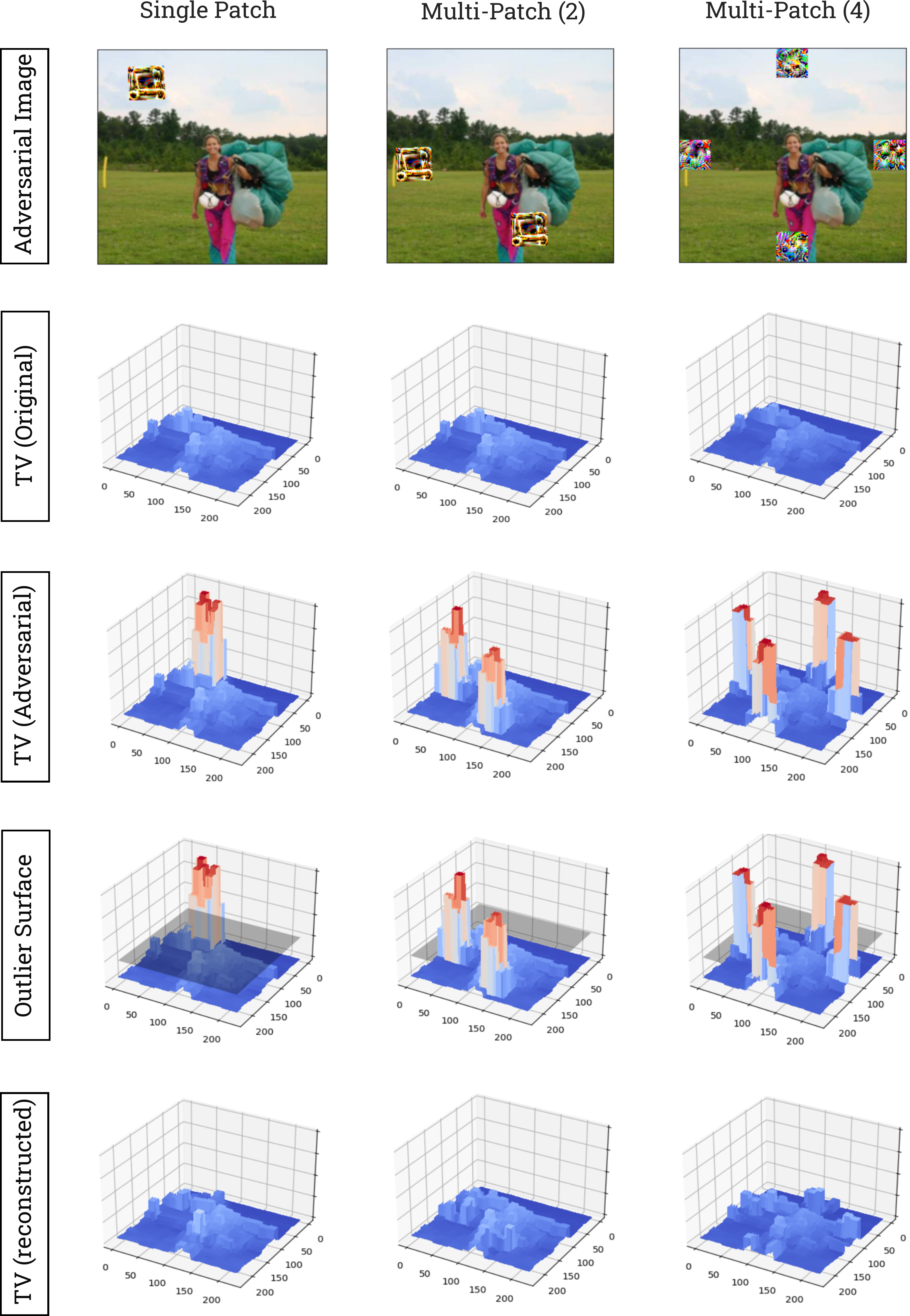}
     \caption{Surface plot of the TV score over the image landscape. The overall size of the patch is between 6-8\% of the image's size.}
     \label{fig:def_tvd}
\end{figure}

This section qualitatively analyzes how TVR works by visualizing 3-D surface plots as shown in Figure \ref{fig:def_tvd}. Although, the TV scores and outliers are calculated channel-wise, for visualization purpose, the 3-D surface plots are made by averaging over three channels. 
The analysis involves testing the patch attack under three scenarios: a  single patch attack, a multi-patch attack using two patches, and a multi-patch attack using four patches. For the first two scenarios, we use a patch with the target class of `Cornet' from the ImageNet-Patch dataset. For the third scenario, we use a multi-patch trained with a `Golf Ball' target class. The original image belongs to the class `Parachute'. 

In the experiment, we visualize the TV score over the image landscape as a surface plot. Hence the x and y axis of the plot is the image's length and width, which is 224 $\times$ 224. We display the TV score of each block against the z-axis. The surface plots consist of a continuous color scale from blue to red, with blue as the lowest score and red as the highest. Figure \ref{fig:def_tvd} shows the plots for four instances: original image, adversarial image, adversarial image with the outlier surface, and finally the reconstructed image. 

In the surface plots, we can see red peaks in the region corresponding to patched areas in the original image. The light gray translucent surface represents the outlier of the TV score for each example. The regions in the surface that are higher than the outlier surface are masked or cropped to zero, which is later reconstructed using a generator. We demonstrate that the TVR defends against multi-patch attacks in the same way it deals with single-patch attacks. The surface plot of a single patch-reconstructed image has the highest resemblances with the original one. Inpainting a cropped image formed post the mitigation of a multi-patch attack is relatively complex. But, we used a simple inpainting generator, and using a better generator will increase the similarity between the original image and the image recovered after a multi-patch attack.

\subsection{Influence of TVR's Block Size on Accuracy}
\label{sec: blk_acc}

The block size is the most critical hyper-parameter affecting the performance of TVR. The outlier in TVR can vary drastically depending on the block size, affecting its defending capability. An appropriate block size depends on the size of the adversarial patch in the scene. The design of most defenses against adversarial attacks requires a conservative estimation of the attack's potency, and TVR is no exception. The attack's potency in the case of digital adversarial attacks like PGD and FGSM is referred to in terms of the noise magnitude. Whereas for patch attacks, patch size proportionally determines the strength of the attack. 

\begin{figure}[t]
\centering
\subfloat[]{
  \centering
  \includegraphics[scale=0.21]{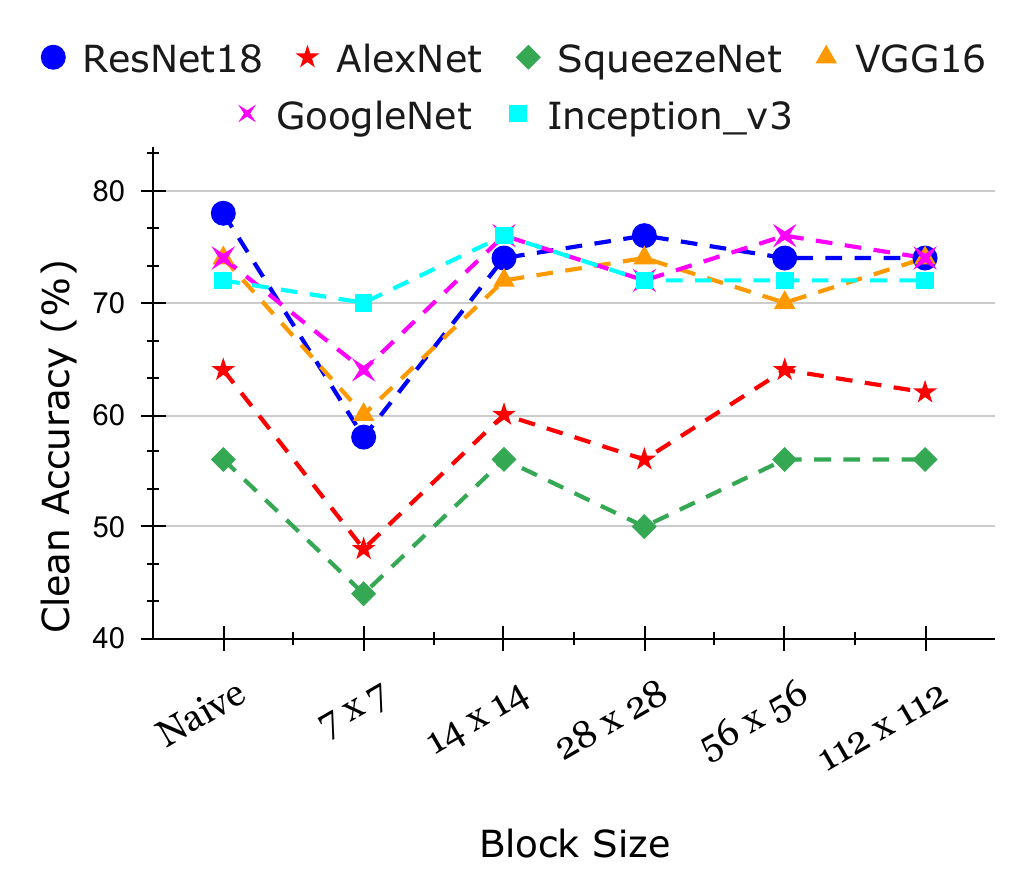}  
  \label{fig:natblk1}
}
\subfloat[]{
  \centering
  \includegraphics[scale=0.21]{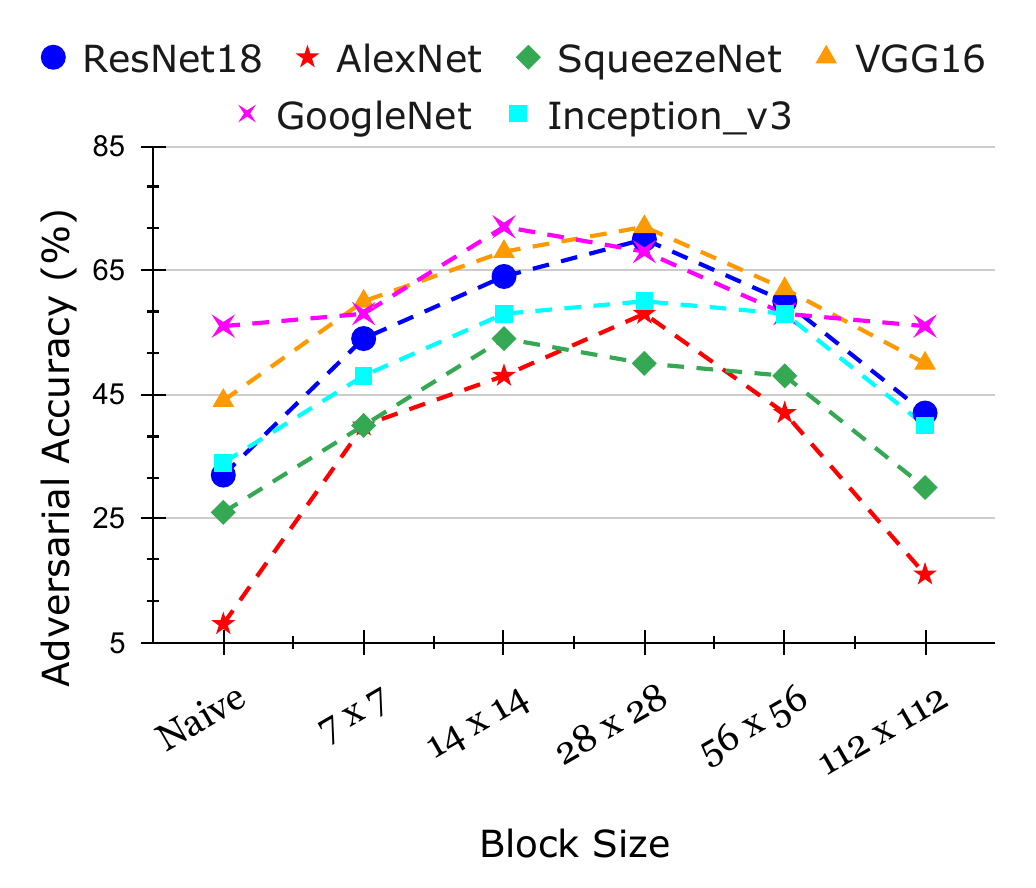}  
  \label{fig:advblk1}
}\\

\subfloat[]{
  \centering
  \includegraphics[scale=0.21]{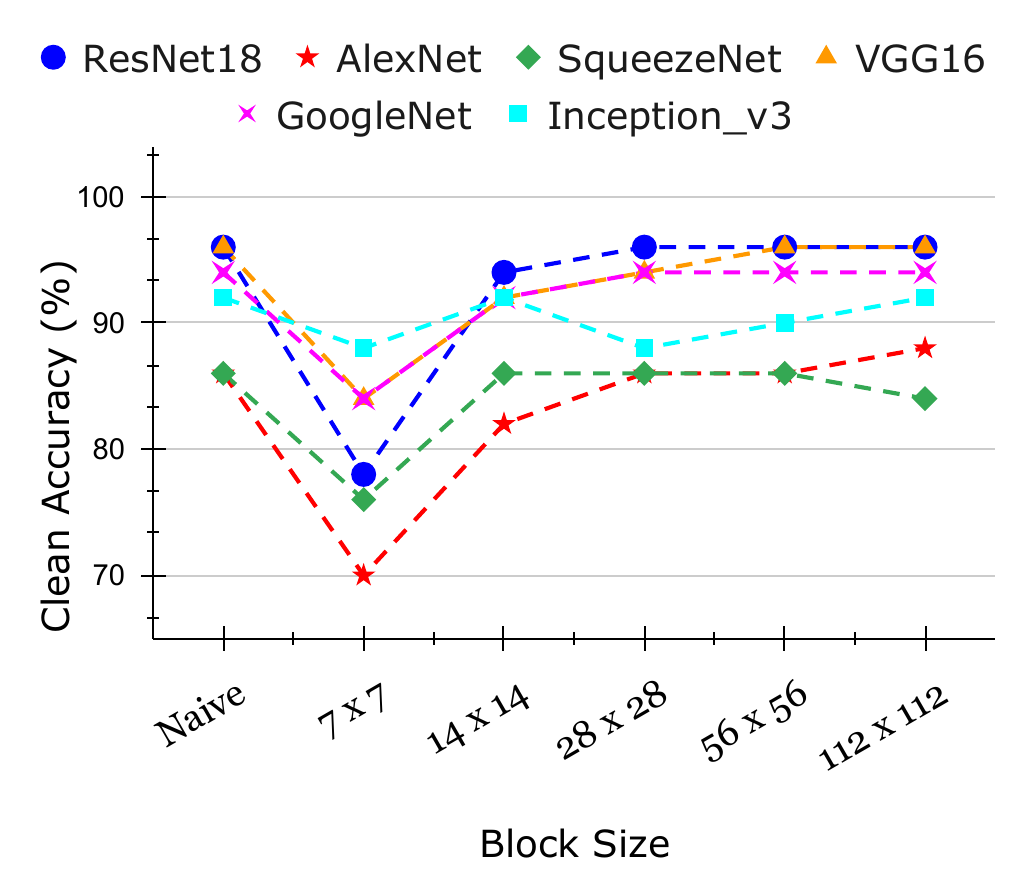}  
  \label{fig:natblk3}
}
\subfloat[]{
  \centering
  \includegraphics[scale=0.21]{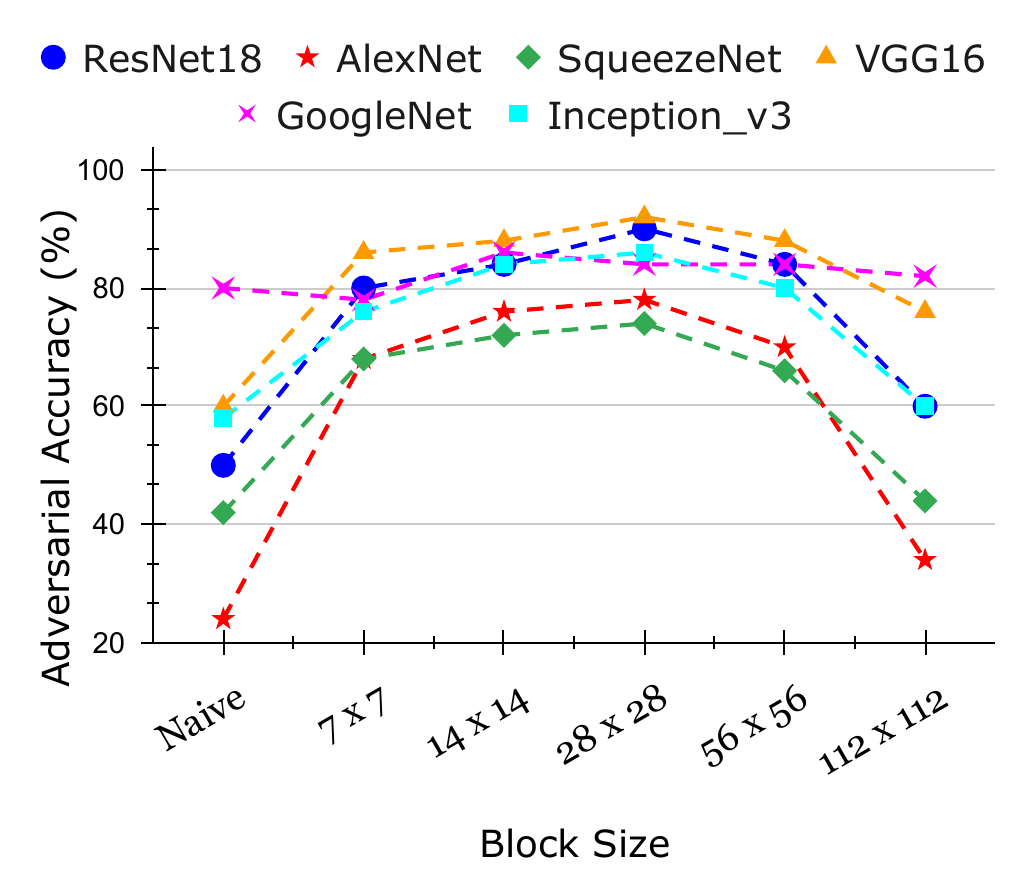}  
  \label{fig:advblk3}
}
\caption{Variation in accuracy with changing TVR's block size. Figure \ref{fig:natblk1} and \ref{fig:advblk1} show clean and adversarial top-1 accuracy. Figure \ref{fig:natblk3} and \ref{fig:advblk3} show natural and adversarial top-3 accuracy.}
\label{fig:blksize}
\end{figure}

In this experiment, we vary the block size from 7 $\times$ 7 to 112 $\times$ 112 by doubling the side length in each step. We evaluate the accuracy of all six models stated in Section \ref{subsec:expsetup} against the `Cornet' adversarial patch (randomly chosen). Figures. As evident from Figure \ref{fig:blksize}, block:7$\times$7 has higher adversarial accuracy than the naive model,but  the natural accuracy decreases considerably. Smaller-sized blocks seem to have a higher tendency for false positives for detecting patches, eventually masking some of the clean regions, which considerably lowers natural accuracy. 

Block:14$\times$14 and block:28$\times$28 have the best performance as they lower the natural accuracy marginally but improve the adversarial accuracy to a large extent. On the other hand, block:112$\times$112 have lower robustness because calculating the total variation score over large regions lowers the chance of detecting patch as outliers. Since top-1 and top-3 adversarial accuracy peaks at block:28$\times$28, which we finally decided as our block size. However, please note that the block size for the best performance will vary based on the conservative assumption of the patch's size.


\begin{figure*}[t]
     \centering 
     \includegraphics[scale=0.135]{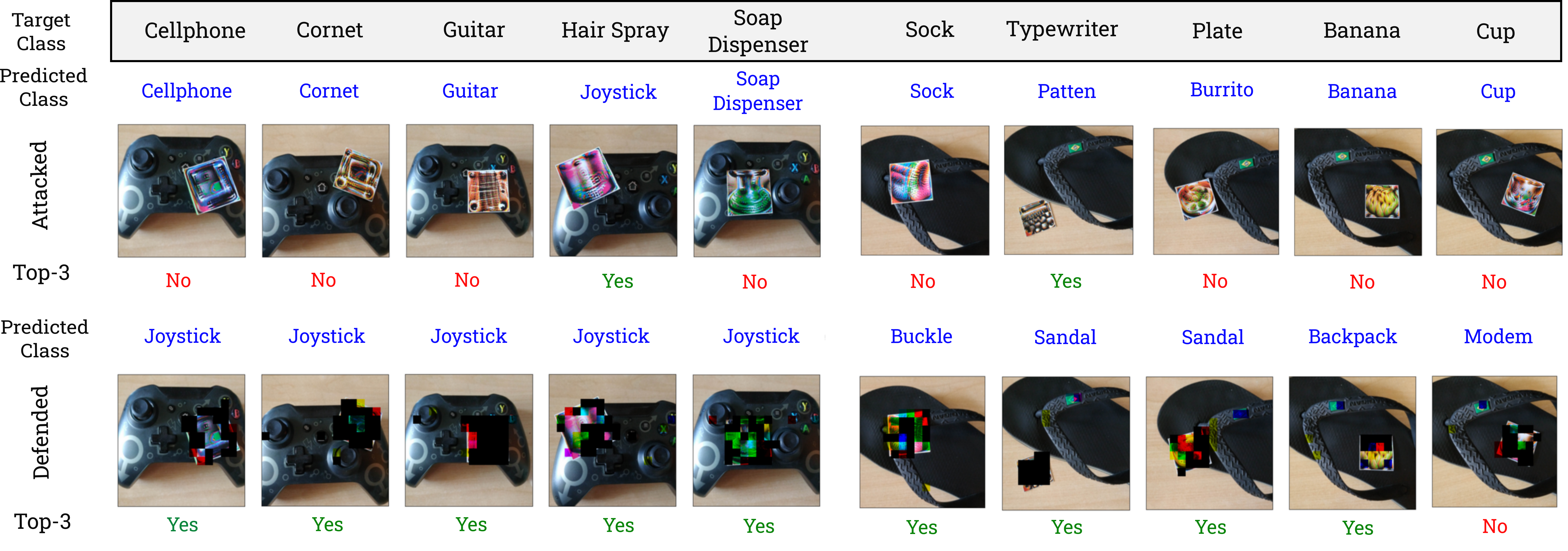}
     \caption{Physical Experiments with Joystick and Sandal}
     \label{fig:phy_exp}
\end{figure*}

\subsection{Influence of Image Inpainting on Accuracy}

As discussed in Section \ref{sec: tvr_form}, the total variation based outlier detection and masking leads to mitigation of the patch attack. Even though TVR nullifies the influence of the patch, we lose some information about the original image. The image inpainting helps to recreate close to an original image using a trained generator. This study evaluates the necessity of image inpainting. We test eleven scenarios (ten patch attacked, and one unattacked) on the ImageNet-Patch benchmark dataset and calculate adversarial accuracy based on images from the test set. 

\begin{figure}[b]
\centering
\includegraphics[scale=0.19]{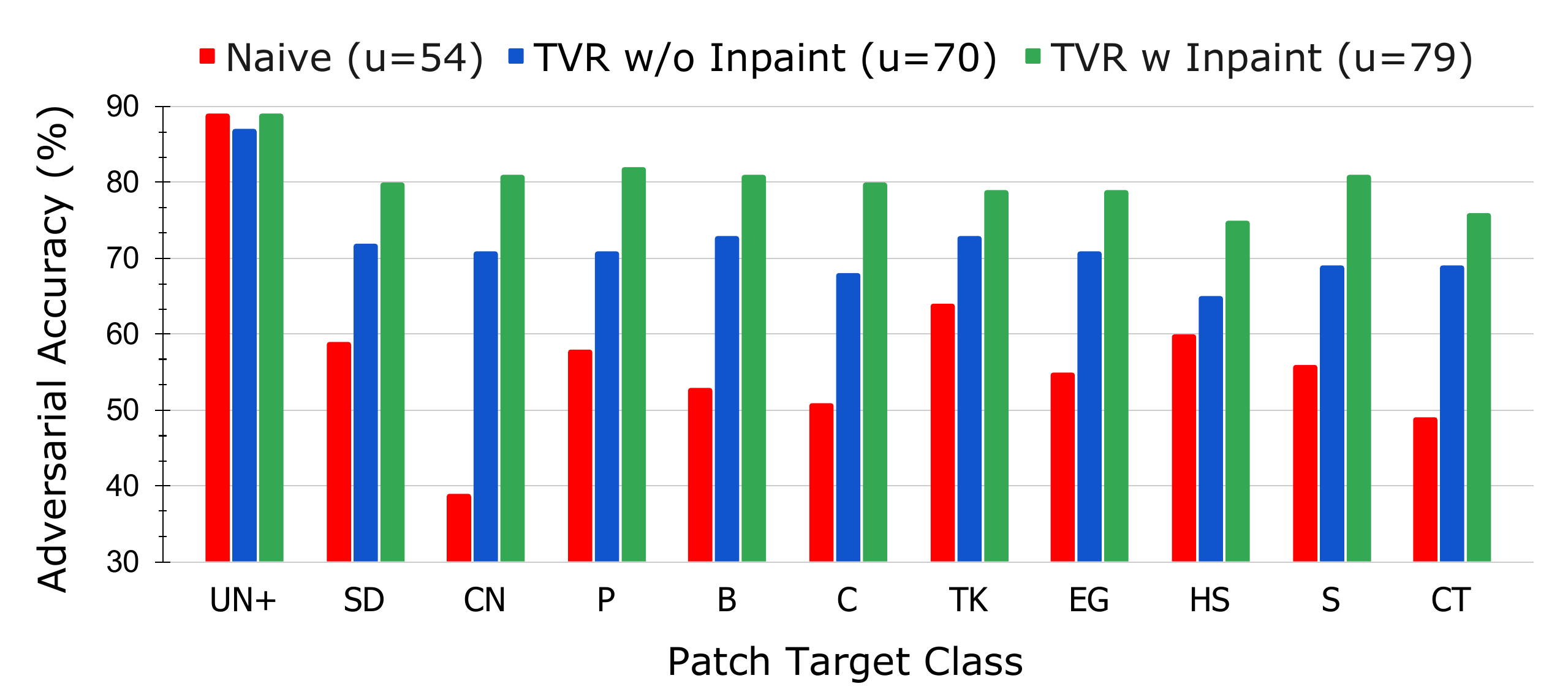}  
\caption{Increment in top-3 accuracy post image inpainting in different scenarios. For patch target class labels, refer Section  \ref{sec: tvr_exp}.
}
\label{fig:inpaint}
\end{figure}

We calculate the accuracy for three scenarios: naive model, TVR without image inpainting, and TVR with image inpainting. Figure \ref{fig:inpaint} show top-3 accuracy highlighting that TVR with image inpainting outperforms the one without against all patches with an improvement of 16\%. We also observe inpainting lowering the gap between natural (unattacked) and adversarial (patch attacked) accuracy. 

\section{Physical Demonstration of TVR}
\label{sec: phy_demo}

A patch attack is a physical world attack, which makes it crucial to test TVR with a real-world example. Validating defensive ability of TVR in the digital setup is insufficient, as the attack may lose potency during its translation from the digital to the physical environment. This study considers two object categories: Joystick and Sandal, as shown in Figure \ref{fig:phy_exp}. These classes also belong to the ImageNet benchmark dataset. The objects are subjected to ten different patches from the ImageNet-Patch dataset. The patches are applied with random transformation and scaling in the range [0.7, 1]. The environmental factors like lighting, which may influence the prediction are maintained. This experiment follows the setup proposed in \cite{pintor2023imagenet}. For this study, we did not perform image inpainting, as our goal is to demonstrate the TVR's ability to mitigate the effect of the patch.

We use a ResNet18 model for the prediction in all test cases. The predicted class shows the top-1 class predicted by the model. We also mention if the predicted class is within the top-3 prediction. As seen in Figure \ref{fig:phy_exp}, we only have 2 out of 10 correct top-3 predictions for the naive model, whereas the correct prediction rose to 9 out of 10 with TVR.  The results show that predicting the adversarial Joystick example is easier. One possible reason might be that Backpack is much closer to Sandal in the feature space (because the straps of the sandals look like the straps of the Backpack). However, there is no such class for the Joystick, making its prediction relatively easier. 

This study can be summarised as follows: First, the size of patches used to attack is larger than that of a digital scenario. It is necessary because the attack efficacy drops while translating the patch from a digital to a physical scenario. Second, the TVR does not have to mask the entire patch to mitigate its malicious influence on the prediction. Instead, it only masks regions with higher TV scores than the outlier of the whole image. Nevertheless, TVR is able to assist the model to make correct predictions in the physical setting. 



\section{Conclusion}

This work proposes a total-variation-based image resurfacing technique to mitigate the threat from adversarial patches. TVR is a model and task-agnostic technique that only depends on the scene. We validate the performance of TVR in both digital and physical environments. The TVR can cleanse single or multiple patches in a single scan of the image. It ensures mitigation of the attack as long as the TV score of perturbations is larger than the outlier threshold. TVR is a first step towards dealing with arbitrary multiple localized perturbations in the scene. 



{\small
\bibliographystyle{ieee_fullname}
\bibliography{egbib}

\begin{thebibliography}{10}\itemsep=-1pt

\bibitem{akhtar2018threat}
Naveed Akhtar and Ajmal Mian.
\newblock Threat of adversarial attacks on deep learning in computer vision: A
  survey.
\newblock {\em Ieee Access}, 6:14410--14430, 2018.

\bibitem{athalye2018synthesizing}
Anish Athalye, Logan Engstrom, Andrew Ilyas, and Kevin Kwok.
\newblock Synthesizing robust adversarial examples.
\newblock In {\em International conference on machine learning}, pages
  284--293. PMLR, 2018.

\bibitem{brown2017adversarial}
Tom~B Brown, Dandelion Man{\'e}, Aurko Roy, Mart{\'\i}n Abadi, and Justin
  Gilmer.
\newblock Adversarial patch.
\newblock {\em arXiv preprint arXiv:1712.09665}, 2017.

\bibitem{chambolle2010introduction}
Antonin Chambolle, Vicent Caselles, Daniel Cremers, Matteo Novaga, and Thomas
  Pock.
\newblock An introduction to total variation for image analysis.
\newblock {\em Theoretical foundations and numerical methods for sparse
  recovery}, 9(263-340):227, 2010.

\bibitem{chen2022towards}
Zhaoyu Chen, Bo Li, Jianghe Xu, Shuang Wu, Shouhong Ding, and Wenqiang Zhang.
\newblock Towards practical certifiable patch defense with vision transformer.
\newblock In {\em Proceedings of the IEEE/CVF Conference on Computer Vision and
  Pattern Recognition}, pages 15148--15158, 2022.

\bibitem{chiang2020certified}
Ping-yeh Chiang, Renkun Ni, Ahmed Abdelkader, Chen Zhu, Christoph Studer, and
  Tom Goldstein.
\newblock Certified defenses for adversarial patches.
\newblock In {\em ICLR}, 2020.

\bibitem{chou2020sentinet}
Edward Chou, Florian Tramer, and Giancarlo Pellegrino.
\newblock Sentinet: Detecting localized universal attacks against deep learning
  systems.
\newblock In {\em 2020 IEEE Security and Privacy Workshops (SPW)}, pages
  48--54. IEEE, 2020.

\bibitem{eykholt2018robust}
Kevin Eykholt, Ivan Evtimov, Earlence Fernandes, Bo Li, Amir Rahmati, Chaowei
  Xiao, Atul Prakash, Tadayoshi Kohno, and Dawn Song.
\newblock Robust physical-world attacks on deep learning visual classification.
\newblock In {\em Proceedings of the IEEE conference on computer vision and
  pattern recognition}, pages 1625--1634, 2018.

\bibitem{gittings2020vax}
Thomas Gittings, Steve Schneider, and John Collomosse.
\newblock Vax-a-net: Training-time defence against adversarial patch attacks.
\newblock In {\em ACCV}, 2020.

\bibitem{guo2017countering}
Chuan Guo, Mayank Rana, Moustapha Cisse, and Laurens Van Der~Maaten.
\newblock Countering adversarial images using input transformations.
\newblock {\em arXiv preprint arXiv:1711.00117}, 2017.

\bibitem{hayes2018visible}
Jamie Hayes.
\newblock On visible adversarial perturbations \& digital watermarking.
\newblock In {\em CVPR Workshops}, pages 1597--1604, 2018.

\bibitem{he2016deep}
Kaiming He, Xiangyu Zhang, Shaoqing Ren, and Jian Sun.
\newblock Deep residual learning for image recognition.
\newblock In {\em Proceedings of the IEEE conference on computer vision and
  pattern recognition}, pages 770--778, 2016.

\bibitem{iandola2016squeezenet}
Forrest~N Iandola, Song Han, Matthew~W Moskewicz, Khalid Ashraf, William~J
  Dally, and Kurt Keutzer.
\newblock Squeezenet: Alexnet-level accuracy with 50x fewer parameters and< 0.5
  mb model size.
\newblock {\em arXiv preprint arXiv:1602.07360}, 2016.

\bibitem{karmon2018lavan}
Danny Karmon, Daniel Zoran, and Yoav Goldberg.
\newblock Lavan: Localized and visible adversarial noise.
\newblock In {\em International Conference on Machine Learning}, pages
  2507--2515. PMLR, 2018.

\bibitem{krizhevsky2017imagenet}
Alex Krizhevsky, Ilya Sutskever, and Geoffrey~E Hinton.
\newblock Imagenet classification with deep convolutional neural networks.
\newblock {\em Communications of the ACM}, 60(6):84--90, 2017.

\bibitem{levine2020randomized}
Alexander Levine and Soheil Feizi.
\newblock ({D}e) randomized smoothing for certifiable defense against patch
  attacks.
\newblock volume~33, pages 6465--6475, 2020.

\bibitem{li2022vip}
Junbo Li, Huan Zhang, and Cihang Xie.
\newblock Vip: Unified certified detection and recovery for patch attack with
  vision transformers.
\newblock In {\em Computer Vision--ECCV 2022: 17th European Conference, Tel
  Aviv, Israel, October 23--27, 2022, Proceedings, Part XXV}, pages 573--587.
  Springer, 2022.

\bibitem{liu2019perceptual}
Aishan Liu, Xianglong Liu, Jiaxin Fan, Yuqing Ma, Anlan Zhang, Huiyuan Xie, and
  Dacheng Tao.
\newblock Perceptual-sensitive gan for generating adversarial patches.
\newblock In {\em Proceedings of the AAAI conference on artificial
  intelligence}, volume~33, pages 1028--1035, 2019.

\bibitem{liu2018dpatch}
Xin Liu, Huanrui Yang, Ziwei Liu, Linghao Song, Hai Li, and Yiran Chen.
\newblock Dpatch: An adversarial patch attack on object detectors.
\newblock {\em arXiv preprint arXiv:1806.02299}, 2018.

\bibitem{mccoyd2020minority}
Michael McCoyd, Won Park, Steven Chen, Neil Shah, Ryan Roggenkemper, Minjune
  Hwang, Jason~Xinyu Liu, and David Wagner.
\newblock Minority reports defense: Defending against adversarial patches.
\newblock In {\em Applied Cryptography and Network Security Workshops: ACNS
  2020 Satellite Workshops, AIBlock, AIHWS, AIoTS, Cloud S\&P, SCI, SecMT, and
  SiMLA, Rome, Italy, October 19--22, 2020, Proceedings}, pages 564--582.
  Springer, 2020.

\bibitem{metzen2021meta}
Jan~Hendrik Metzen, Nicole Finnie, and Robin Hutmacher.
\newblock Meta adversarial training against universal patches.
\newblock {\em arXiv preprint arXiv:2101.11453}, 2021.

\bibitem{mnih2015human}
Volodymyr Mnih, Koray Kavukcuoglu, David Silver, Andrei~A Rusu, Joel Veness,
  Marc~G Bellemare, Alex Graves, Martin Riedmiller, Andreas~K Fidjeland, Georg
  Ostrovski, et~al.
\newblock Human-level control through deep reinforcement learning.
\newblock {\em nature}, 518(7540):529--533, 2015.

\bibitem{naseer2019local}
Muzammal Naseer, Salman Khan, and Fatih Porikli.
\newblock Local gradients smoothing: Defense against localized adversarial
  attacks.
\newblock In {\em 2019 IEEE Winter Conference on Applications of Computer
  Vision (WACV)}, pages 1300--1307. IEEE, 2019.

\bibitem{pintor2023imagenet}
Maura Pintor, Daniele Angioni, Angelo Sotgiu, Luca Demetrio, Ambra Demontis,
  Battista Biggio, and Fabio Roli.
\newblock Imagenet-patch: A dataset for benchmarking machine learning
  robustness against adversarial patches.
\newblock {\em Pattern Recognition}, 134:109064, 2023.

\bibitem{rao2020adversarial}
Sukrut Rao, David Stutz, and Bernt Schiele.
\newblock Adversarial training against location-optimized adversarial patches.
\newblock In {\em Computer Vision--ECCV 2020 Workshops: Glasgow, UK, August
  23--28, 2020, Proceedings, Part V 16}, pages 429--448. Springer, 2020.

\bibitem{rebuffi2021data}
Sylvestre-Alvise Rebuffi, Sven Gowal, Dan~Andrei Calian, Florian Stimberg,
  Olivia Wiles, and Timothy~A Mann.
\newblock Data augmentation can improve robustness.
\newblock {\em Advances in Neural Information Processing Systems},
  34:29935--29948, 2021.

\bibitem{ren2021adversarial}
Huali Ren, Teng Huang, and Hongyang Yan.
\newblock Adversarial examples: attacks and defenses in the physical world.
\newblock {\em International Journal of Machine Learning and Cybernetics},
  pages 1--12, 2021.

\bibitem{salman2022certified}
Hadi Salman, Saachi Jain, Eric Wong, and Aleksander Madry.
\newblock Certified patch robustness via smoothed vision transformers.
\newblock In {\em Proceedings of the IEEE/CVF Conference on Computer Vision and
  Pattern Recognition}, pages 15137--15147, 2022.

\bibitem{sharif2016accessorize}
Mahmood Sharif, Sruti Bhagavatula, Lujo Bauer, and Michael~K Reiter.
\newblock Accessorize to a crime: Real and stealthy attacks on state-of-the-art
  face recognition.
\newblock In {\em Proceedings of the 2016 acm sigsac conference on computer and
  communications security}, pages 1528--1540, 2016.

\bibitem{sharma2022adversarial}
Abhijith Sharma, Yijun Bian, Phil Munz, and Apurva Narayan.
\newblock Adversarial patch attacks and defences in vision-based tasks: A
  survey.
\newblock {\em arXiv preprint arXiv:2206.08304}, 2022.

\bibitem{sharma2023vulnerability}
Abhijith Sharma, Yijun Bian, Vatsal Nanda, Phil Munz, and Apurva Narayan.
\newblock Vulnerability of cnns against multi-patch attacks.
\newblock In {\em Proceedings of the 2023 ACM Workshop on Secure and
  Trustworthy Cyber-Physical Systems}, pages 23--32, 2023.

\bibitem{shorten2019survey}
Connor Shorten and Taghi~M Khoshgoftaar.
\newblock A survey on image data augmentation for deep learning.
\newblock {\em Journal of big data}, 6(1):1--48, 2019.

\bibitem{siddiqui2021novel}
Abdul~Jabbar Siddiqui and Azzedine Boukerche.
\newblock A novel lightweight defense method against adversarial patches-based
  attacks on automated vehicle make and model recognition systems.
\newblock {\em Journal of Network and Systems Management}, 29(4):41, 2021.

\bibitem{simonyan2014very}
Karen Simonyan and Andrew Zisserman.
\newblock Very deep convolutional networks for large-scale image recognition.
\newblock {\em arXiv preprint arXiv:1409.1556}, 2014.

\bibitem{szegedy2015going}
Christian Szegedy, Wei Liu, Yangqing Jia, Pierre Sermanet, Scott Reed, Dragomir
  Anguelov, Dumitru Erhan, Vincent Vanhoucke, and Andrew Rabinovich.
\newblock Going deeper with convolutions.
\newblock In {\em Proceedings of the IEEE conference on computer vision and
  pattern recognition}, pages 1--9, 2015.

\bibitem{szegedy2016rethinking}
Christian Szegedy, Vincent Vanhoucke, Sergey Ioffe, Jon Shlens, and Zbigniew
  Wojna.
\newblock Rethinking the inception architecture for computer vision.
\newblock In {\em Proceedings of the IEEE conference on computer vision and
  pattern recognition}, pages 2818--2826, 2016.

\bibitem{taigman5closing}
Y Taigman, M Yang, M Ranzato, and L Wolf.
\newblock Closing the gap to human-level performance in face verification.
  deepface.
\newblock In {\em Proceedings of the IEEE Computer Vision and Pattern
  Recognition (CVPR)}, volume~5, page~6.

\bibitem{wei2022physical}
Hui Wei, Hao Tang, Xuemei Jia, Hanxun Yu, Zhubo Li, Zhixiang Wang, Shin'ichi
  Satoh, and Zheng Wang.
\newblock Physical adversarial attack meets computer vision: A decade survey.
\newblock {\em arXiv preprint arXiv:2209.15179}, 2022.

\bibitem{xiang2020patchguard}
Chong Xiang, Arjun~Nitin Bhagoji, Vikash Sehwag, and Prateek Mittal.
\newblock Patchguard: Provable defense against adversarial patches using masks
  on small receptive fields.
\newblock {\em arXiv preprint arXiv:2005.10884}, 2020.

\bibitem{xiang2022patchcleanser}
Chong Xiang, Saeed Mahloujifar, and Prateek Mittal.
\newblock $\{$PatchCleanser$\}$: Certifiably robust defense against adversarial
  patches for any image classifier.
\newblock In {\em 31st USENIX Security Symposium (USENIX Security 22)}, pages
  2065--2082, 2022.

\bibitem{xiang2021patchguard++}
Chong Xiang and Prateek Mittal.
\newblock Patchguard++: Efficient provable attack detection against adversarial
  patches.
\newblock {\em arXiv preprint arXiv:2104.12609}, 2021.

\end{thebibliography}
}

\end{document}